\documentclass{article}
\usepackage{spconf,amsmath,graphicx}
\usepackage{multirow}

\title{Exponential Moving Average Model in Parallel Speech Recognition Training}
\name{Xu Tian, Jun Zhang, Zejun Ma, Yi He, Juan Wei}

\address{Alibaba Shenma Search\\
	\small \texttt\{xu.tian, zj102217, zejun.mamzj, heyi.hy, wj80290\}@alibaba-inc.com}

\begin{document}

\maketitle
\begin{abstract}
As training data rapid growth, large-scale parallel training with multi-GPUs cluster is widely applied in the neural network model learning currently.
We present a new approach that applies exponential moving average method in large-scale parallel training of neural network model. It is a non-interference strategy that the exponential moving average model is not broadcasted to distributed workers to update their local models after model synchronization in the training process, and it is implemented as the final model of the training system. Fully-connected feed-forward neural networks (DNNs) and deep unidirectional Long short-term memory (LSTM) recurrent neural networks (RNNs) are successfully trained with proposed method for large vocabulary continuous speech recognition on Shenma voice search data in Mandarin. The character error rate (CER) of Mandarin speech recognition further degrades than state-of-the-art approaches of parallel training.
\end{abstract}


\section{Introduction}
\label{sec:intro}
Over the past few years, neural networks has been widely used in some domains, such as large vocabulary continuous speech recognition (LVCSR) \cite{graves2014towards,graves2013hybrid,amodei2015deep}, image recognition \cite{krizhevsky2012imagenet,he2016deep} and neural machine translation \cite{bahdanau2014neural}.
Fully-connected feed-forward deep neural networks (DNNs) and Recurrent neural networks (RNNs), especially long short-term memory (LSTM) RNNs have shown the effective performance for LVCSR \cite{hinton2012deep,graves2013speech,sak2014long,zhang2016highway}.
It is significant that training with larger dataset could improve recognition accuracy. As a matter of fact, larger dataset does mean more training samples and more model parameters, and it is high time consumption to train neural networks with only one computing unit. Therefore, parallel training with multi-GPUs is essential, but it leads to slower convergence. For multi-GPUs training, the key problem is how to accelerate convergence and get further improvement. 

Mini-batch based stochastic gradient descent (SGD) is the most prevalent method in neural network training procedure. Several methods are proposed based on it, and achieving encouraging performance for parallel training.
Asynchronous SGD is a successful attempt \cite{dean2012large,zhang2013asynchronous}, and it is shown that parallel training with asynchronous SGD can many times speedup without lowering the accuracy.
Besides, synchronous SGD is another positive effort, where the parameter server waits for every workers to finish their computation and send their local models to it, and then it sends updated model back to all workers \cite{chen2016revisiting}. Synchronous SGD converges well in parallel training with data parallelism, and is also easy to  be implemented.

Model averaging is a method for large-scale parallel training, which the final model is averaged from all parameters of separated models \cite{mcdonald2010distributed,zinkevich2010parallelized}. Compared with single GPU training, it achieves linear speedup, but the accuracy decreases. Moreover, blockwise model-updating filter (BMUF) provides another linear speedup approach with multi-GPUs on the basis of model averaging. It can achieve improvement or no-degradation of recognition performance compared with mini-batch SGD on single GPU \cite{chen2016scalable}.

It is demonstrated that the performance of moving average of the parameters obtained by SGD is as good as that of the parameters which minimize the empirical cost, 
and moving average parameters can be used as the estimator of them, if there are enough training samples \cite{polyak1992acceleration}. 
One pass learning is then proposed, and it is the combination of learning rate schedule and averaged SGD using moving average \cite{xu2011towards}. When the moving average model outperforms the model aggregated with model averaging, the moving average model is broadcasted to update local workers. Since the learning rate of one pass learning is difficult to be adjusted, it is challenging to train different models in different domains.

In this paper, we propose a new approach which applies exponential moving average (EMA) directly in large-scale synchronous-based parallel training. It is a kind of non-interference method that the EMA model is not broadcasted, after the parameters of each worker are synchronized. It is applied as the final model of the training. The exponential moving average method in parallel training will be described in Section~\ref{sec:ema}. Neural network models are successfully trained for LVCSR, using this method. The experiments and results are presented in Section~\ref{sec:exp}, followed by the conclusion in Section~\ref{sec:conclusion}.

\section{Exponential Moving Average Model}
\label{sec:ema}

In recognition applications, the parameters $\theta$ of neural network is trained for classification. It's also an optimization problem:
$$arg \min_\theta\frac{1}{t}\sum_{i=1}^t(L(f_\theta(x),y))$$
Where $t$ is the number of data points, $(x,y)$ is the input data and correspondent target, $L$ is the loss function, and $f_\theta$ denotes the network. Let $\theta^*$ be the parameters that minimize the empirical cost. Large scale recognition training needs to deal with the optimization problem with billions of training data, and makes it hard to find the  $\theta^*$. SGD and its variants are presented promising learning results for large scale optimization problem, and become the most popular methods of deep learning.

\subsection{Model averaging and block-wise model updating filter}

In order to reduce the time cost of training, data parallelism is implemented. The full training dataset is partitioned into $N$ splits without overlapping, and they are distributed to $N$ GPUs. 

Each GPU optimizes local model in parallel with one split of training dataset. 
After a mini-batch training, the global model is needed to update, and it is computed with model averaging or BMUF, and consequently broadcasted to GPUs to update their local models. 
For model averaging method, all local models are synchronized and averaged, and then aggregated model $\bar \theta(t)$ is sent back to GPUs \cite{mcdonald2010distributed,zinkevich2010parallelized}.
For BMUF method, the global model $\theta_g(t)$ is employed, instead of $\bar \theta(t)$ in model averaging method.
The synchronization and updating process of $\theta_g(t)$ in BMUF as follows:
$$\bar\theta(t)=\frac{1}{N}\sum^N_{i=1}\theta_i$$
$$G(t)= \bar \theta(t) - \theta_g(t-1)​$$
$$\Delta(t)=\eta_t\Delta(t-1)+\zeta_tG(t)$$
Where $G(t)$ denotes model update, and $\Delta(t)$ is the global-model update. There are two parameters in BMUF, block momentum $\eta$, and block learning rate $\zeta$. Then, the global model is updated as 
$$\theta_g(t)=\theta_g(t-1)+\Delta(t)$$
Consequently, $\theta_g(t)$ is broadcasted to all GPUs to update their local models.

It is worth noting that when block momentum and block learning rate are set as 0 and 1, BMUF becomes model averaging. We treat model averaging and BMUF as model averaging based methods.

\subsection{Moving Average and Exponential Moving Average}
Averaged SGD is proposed to further accelerate the convergence speed of SGD. Averaged SGD leverages the moving average (MA) $\bar\theta$ as the estimator of $\theta^*$ \cite{polyak1992acceleration}:
$$\bar{\theta_t} = \frac{1}{t}\sum_{\tau=1}^{t}\theta_\tau$$ 
Where $\theta_{\tau}$ is computed by model averaging or BMUF. It is shown that $\bar\theta_t$ can well converge to $\theta^*$, with the large enough training dataset in single GPU training. It can be considered as a non-interference strategy that $\bar\theta_t$ does not participate the main optimization process, and only takes effect after the end of entire optimization. 
However, for the parallel training implementation, each $\theta_\tau$ is computed by model averaging and BMUF with multiple models, and moving average model $\bar\theta_t$ does not well converge, compared with single GPU training.

Model averaging based methods are employed in parallel training of large scale dataset, because of their faster convergence, and especially no-degradation implementation of BMUF. But combination of model averaged based methods and moving average does not match the expectation of further enhance performance and it is presented as
$$\bar\theta_{g_t} = \frac{1}{t}\sum_{\tau=1}^{t}\theta_{g_\tau}$$ 
The weight of each $\theta_{g_t}$ is equal in moving average method regardless the effect of temporal order. But $t$ closer to the end of training achieve higher accuracy in the model averaging based approach, and thus it should be with more proportion in final $\bar \theta_g$. As a result, exponential moving average(EMA) is appropriate, which the weight for each older parameters decrease exponentially, and never reaching zero. 
After moving average based methods, the EMA parameters are updated recursively as
$$\hat \theta_{g_t}=\alpha\hat\theta_{g_{t-1}}+(1-\alpha)\theta_{g_t}$$
Here $\alpha$ represents the degree of weight decrease, and called exponential updating rate. 
EMA is also a non-interference training strategy that is implemented easily, as the updated model is not broadcasted. Therefore, there is no need to add extra learning rate updating approach, as it can be appended to existing training procedure directly.

\section{Experiments and Results}
\label{sec:exp}
\subsection{Training Data}

In order to present the performance of our proposed method, we trained acoustic model for LVCSR. A large quantity of labeled data is needed for training a more accurate acoustic model. We collect the 17000 hours labeled data from Shenma voice search, which is one of the most popular mobile search engines in China. The dataset is created from anonymous online users' search queries in Mandarin, and all audio file's sampling rate is 16kHz, recorded by mobile phones. This dataset consists of many different conditions, such as diverse noise even low signal-to-noise, babble, dialects, accents, hesitation and so on. The dataset is divided into training set, validation set and test set, and the quantity of them is shown in Table~\ref{table:data}. 
The three sets are split according to speakers, in order to avoid utterances of same speaker appearing in three sets simultaneously.
The overfitting can be prevented in time, if there is a apparent gap between the frame error rate (FER) of training and validation set.

\begin{table}[t]
	\centering
	
	\begin{tabular}{|c|c|}
		\hline
		Dataset & Hours  \\ 
		\hline
		\hline
		Training set  & 16150  \\
		\hline
		Validation set & 850   \\
		\hline
		Test set &  10 \\
		\hline
		\bf Total & 17010 \\
		\hline
	\end{tabular}	
	\caption{ \label{table:data} \small The time summation of different sets.}
\end{table}


\subsection{Experimental setup}
LSTM RNNs outperform conventional RNNs for speech recognition system, especially deep LSTM RNNs, because of its long-range dependencies more accurately for temporal sequence conditions \cite{hermans2013training,sak2015acoustic}. Shenma voice search is a streaming service that intermediate recognition results displayed while users are still speaking. So as for online recognition in real time, we prefer unidirectional LSTM model rather than bidirectional one. Thus, the parallel training procedure is unidirectional LSTM-based.

A 28-dimensional filter bank feature is extracted for each frame, and is concatenated with first and second order difference as the final input of the network. The architecture we trained consists of two LSTM layers with sigmoid activation function, followed by a full-connection layer. The out layer is a softmax layer with 11088 hidden markov model (HMM) tied-states as output classes, the loss function is cross-entropy (CE). The performance metric of the system in Mandarin is reported with character error rate (CER). The alignment of frame-level ground truth is obtained by GMM-HMM system. Mini-batched SGD is utilized with momentum trick and the network is trained for a total of 4 epochs. The block learning rate and block momentum of BMUF are set as 1 and 0.9. 5-gram language model is leveraged in decoder, and the vocabulary size is as large as 760000.

EMA method is proposed for parallel training problem. In our training system, it is employed on the MPI-based HPC cluster where 8 GPUs are used to train neural network models. Each GPU processes non-overlap subset split from the entire large scale dataset in parallel. 

Local models from distributed workers synchronize with each other in decentralized way. In the traditional model averaging and BMUF method, a parameter server waits for all workers to send their local models, aggregate them, and send the updated model to all workers. Computing resource of workers is wasted until aggregation of the parameter server done. Decentralized method makes full use of computing resource. There is no centralized parameter server, and peer to peer communication is used to transmit local models between workers. Local model $\theta_i$ of $i$-th worker in $N$ workers cluster is split to $N$ pieces $\theta_{i,j}\ j=1\cdots N$, and send to corresponding worker. In the aggregation phase, $j$-th worker computed $N$ splits of model $\theta_{i,j}\ i=1\cdots N$ and send updated model $\bar \theta_{g_j}$ back to workers. As a result, all workers participate in aggregation and no computing resource is dissipated. It is significant to promote training efficiency, when the size of neural network model is too large. The EMA model is also updated additionally, but not broadcasting it.

Besides, frame stacking leads to reduce the computation and training time dramatically~\cite{sak2015fast}. Frames are stacked so as that the network sees multiple frames at a time. The super frame after stacking is the input feature of the network and it contains abundant information. As a result, 3 frames are stacked without overlapping in the training procedure. 

\subsection{Results}
The test set including about 9000 samples contains various real world conditions. It simulates the majority of user scenarios, and could well evaluate the performance of a trained model. BMUF based approach, which has no worse performance than the single-GPU training procedure, is the baseline of experiments. The results of MA and EMA methods on the basis of BMUF are presented, and we call them MA-based methods.

\begin{figure}[ht]
	\vskip 0.2in
	\begin{center}
		\includegraphics[width=\columnwidth]{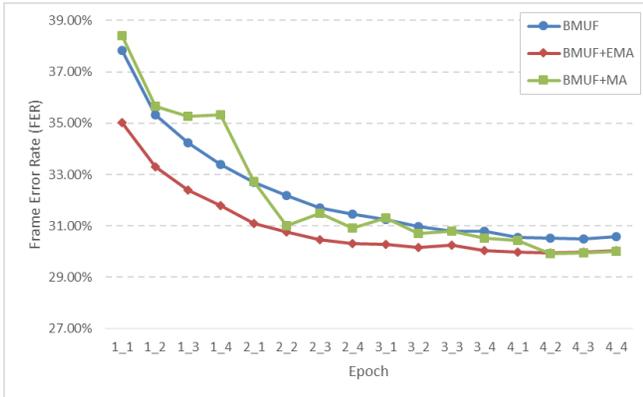}
		\caption{Curves of FER on validation set with different methods.}
		\label{fig:facurve}
	\end{center}
	\vskip -0.2in
\end{figure}

\begin{figure}[ht]
	\vskip 0.2in
	\begin{center}
		\includegraphics[width=\columnwidth]{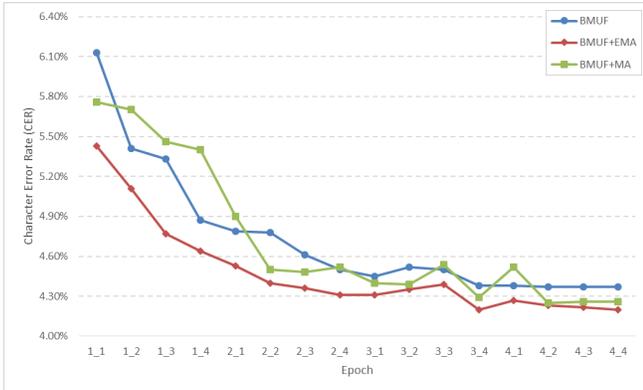}
			\caption{Curves of CER on test set with different methods, and the model is extracted from the LSTM training process.}
			\label{fig:cercurve}
	\end{center}
		\vskip -0.2in
\end{figure}

Since the EMA is a non-interference method, the performance can not be evaluated with the real-time FER. Therefore, the FER on validation sets are computed after every epoch. In order to present the decoding performance of MA-based methods, we extract 4 temporary models from each epoch to visualize the degradation of CER. FER curves of LSTM models trained with BMUF, MA and EMA methods are shown in Figure~\ref{fig:facurve}. It is significant that the frame accuracy of MA-based methods are higher than those of BMUF. Frame accuracies between MA-based methods have slight difference. Though the EMA only perform better than MA slightly on FER, there is an obvious difference between the CER of EMA and MA, as shown in Figure~\ref{fig:cercurve} which illustrates CER curves of different models after decoding. It demonstrates that decoding result of EMA is always much better than that of BMUF, but that of MA fluctuates greatly, and even higher than that of BMUF sometimes.
From the Table~\ref{table:cer} which shows the CER of final models trained from three methods, the superiority of EMA over the others can be also observed. EMA method achieves about relative 3.9\% CER reduction on test set, while MA method only achieves relative 2.1\% CER reduction. 

Moreover, the CER of final DNN models with 8 layers are also presented in Table~\ref{table:cer}, and the CER of EMA method decreases relative 8.4\% compared with baseline.
Therefore, more accuracy models are trained with large-scale parallel training using EMA method, and it is more stable than MA method.


\begin{table}[t]
	\centering	
	\noindent
	\renewcommand{\multirowsetup}{\centering}  
	\begin{tabular}{|c|c|c|} 
		\hline  
		\multirow{2}{1cm}{Methods} & \multicolumn{2}{|c|}{CER (\%)} \\ \cline{2-3}
		& LSTM & DNN \\ 
		\hline
		\hline
		BMUF  & 4.37  & 7.52 \\
		\hline
		BMUF + MA & 4.26  & 7.14 \\
		\hline
		BMUF + EMA & 4.20  & 6.89  \\
		\hline
	\end{tabular}
	\caption{ \label{table:cer} \small The CER of LSTM and DNN final models trained with three methods.}
\end{table}



\section{Conclusion}
\label{sec:conclusion}

The exponential moving average method is proposed in this paper for multi-GPUs cluster parallel training with almost linear speedup. It is demonstrated that unidirectional LSTM and DNN models trained with EMA method have better decoding results than that of BMUF and traditional moving average methods for large vocabulary continues speech recognition in Mandarin. Our future work includes 1) Employing this method to CNN, Connectionist Temporal Classification (CTC), attention-based neural networks and other hybrid deep neural network architecture; 2) Extending this method from frame-wise discriminative training to sequence discriminative training such as maximum mutual information (MMI) and segmental Minimum Bayes-Risk (sMBR); 3) Develop more approaches for parallel training with better performance.


\bibliographystyle{IEEEbib}
\bibliography{ref}

\end{document}